\documentclass[letterpaper, 10 pt, conference]{ieeeconf}
\IEEEoverridecommandlockouts
\usepackage{cite}
\usepackage{amsmath,amssymb,amsfonts}
\usepackage{algorithmic}
\usepackage{graphicx}
\usepackage{textcomp}
\usepackage{xcolor}
\usepackage{amsmath}
\usepackage{booktabs}
\usepackage{makecell}
\usepackage{cancel}
\usepackage[normalem]{ulem}
\usepackage{etoolbox}
\def\BibTeX{{\rm B\kern-.05em{\sc i\kern-.025em b}\kern-.08em
    T\kern-.1667em\lower.7ex\hbox{E}\kern-.125emX}}

\newcommand{\nip}[1]{\par\addvspace{1ex}\noindent\textbf{#1}}

\newcommand\blfootnote[1]{%
  \begingroup
  \renewcommand\thefootnote{}\footnote{#1}%
  \addtocounter{footnote}{-1}%
  \endgroup
}

\begin{document}

\title{\LARGE \bf Backdoors in Learning-Based Industrial Robotic Arm Manipulation: An Empirical Security Study\\
}

\author{%
\ifdef{\latexml}{%
  Zijian Zhang$^{1}$, Zhen Zeng$^{1}$ (Senior Member, IEEE), Zhongshu Gu$^{2}$ and Sandeep Pisharody$^{3}$%
}{%
  Zijian Zhang$^{1}$, Zhen Zeng$^{1}$, \textit{Senior Member}, IEEE, Zhongshu Gu$^{2}$ and Sandeep Pisharody$^{3}$%
}%
\thanks{$^{1}$Zijian Zhang and Zhen Zeng are with the Department of Computer Science, 
        University of Wisconsin Milwaukee, Milwaukee, WI, USA, 
        {\tt\small zhang474, zhenzeng@uwm.edu}}%
\thanks{$^{2}$Zhongshu Gu is with IBM Research,
        Yorktown Heights, NY, USA,
        {\tt\small zgu@us.ibm.com}}%
\thanks{$^{3}$Sandeep Pisharody is with MIT Lincoln Laboratory,
        Lexington, MA, USA,
        {\tt\small sandeep.pisharody@ll.mit.edu}}%
}

\maketitle

\blfootnote{Presented at the IEEE/RSJ International Conference on Intelligent Robots and Systems (IROS 2026) Workshop on Industrial Applications for Robot Learning, 2026, Pittsburgh, PA, USA}

\blfootnote{DISTRIBUTION STATEMENT A. Approved for public release. Distribution is unlimited. This material is based upon work supported by the Department of the Air Force under Air Force Contract No. FA8702-15-D-0001 or FA8702-25-D-B002. Any opinions, findings, conclusions or recommendations expressed in this material are those of the author(s) and do not necessarily reflect the views of the Department of the Air Force.}

\newcommand{\zz}[1]{\textcolor{magenta}{\{zz: #1\}}}

\newcommand{\zg}[1]{\textcolor{purple}{\{zg: #1\}}}

\newcommand{\zj}[1]{\textcolor{commentgreen}{\{zj: #1\}}}
\newcommand{\revzj}[1]{\textcolor{commentgreen}{\{#1\}}}

\newcommand{\ssp}[1]{\textcolor{blue}{\{ssp: #1\}}}

\newcommand{\rev}[1]{\textcolor{red}{\{#1\}}}

\definecolor{commentgreen}{RGB}{0,110,70}
\definecolor{customcyan}{HTML}{d0e1eb}
\definecolor{yellow}{HTML}{fbd55b}

\newcommand{\circnum}[1]{%
\tikz[baseline=(n.base)]{
\node[
circle,
fill=black,
text=white,
font=\bfseries\scriptsize,
inner sep=1.2pt
] (n) {#1};
}}

\begin{abstract}
Learning-based models (e.g., visuomotor and Vision-Language-Action (VLA)) are increasingly explored for industrial robotic manipulation, where model predictions are directly translated into physical actions. This tight coupling between model behavior and physical execution makes hidden security vulnerabilities particularly consequential. 
While backdoor attacks have been widely studied in conventional AI models, their effects on deployed learning-based robotic arm manipulation systems remain less understood: a backdoored robot can behave normally during benign operation while inducing attacker-specified behaviors only when specific triggers are present, posing potentially serious risks in physical environments. 

In this work, we present a preliminary empirical security study of backdoor attacks and defenses in learning-based robotic manipulation on two real commercial industrial robotic arms (FANUC and xArm). We investigate whether a backdoor can reliably induce semantically incorrect manipulation behaviors while remaining stealthy under nominal task execution. We then develop an online defense pipeline that detects and neutralizes triggers at runtime, and compare its effectiveness against an offline fine-tuning defense. Beyond defense effectiveness, we further evaluate the computational latency and execution overhead introduced by the defense pipeline to assess its suitability for high-throughput industrial operation.

\end{abstract}


\section{Introduction}
\label{sec:introduction}

Driven by advances in vision-language foundation models and visuomotor learning, modern industrial manipulators increasingly adopt data-driven architectures that map high-dimensional visual observations directly to robot actions~\cite{black2410pi0,shukor2025smolvla,bjorck2025gr00t_n1,chi2023diffusion,zhao2023learning}. As these policies move from controlled laboratory settings into physical production environments, the stakes for operational safety become even higher. 
This transition to learning-based planning introduces a fundamentally new attack surface through the \textit{AI model supply chain}, where training datasets, demonstrations, or pretrained models may originate from third parties and cannot always be fully trusted. A particularly concerning threat is the backdoor attack, in which
an adversary can poison third-party demonstration datasets to \textit{intentionally} alter policy behavior. A backdoored robot can behave normally during benign operation while introducing attacker-specified behaviors only when specific triggers are present. While backdoor attacks and defenses have been widely studied in conventional AI models~\cite{dattabackdoors_agent_datapoisoning_outsourced,yan2024llm_code_generation,zhai2023text_image_generation,zhang2024instruction_text_generation}, their implications for physical robotic arm manipulation remain largely underexplored, particularly on real industrial arm.

We outline the key challenges that have limited security studies of backdoors in such settings from both the \textbf{attack} and \textbf{defense} perspectives. \textit{Conducting effective backdoor attacks on industrial robotic arms requires addressing two key challenges:} 1) constructing a semantically meaningful and stealthy backdoor behavior. Robotic policies generate continuous action trajectories, and a successful backdoor must induce a coherent task-level failure through these temporally correlated actions. Conducting such attacks requires one to identify and modify a consistent manipulation segment so that the poisoned behavior expresses a distinct malicious intent while preserving the surrounding nominal execution.
2) ensuring physical feasibility of malicious behavior in robotic arms. A backdoor may induce the intended semantic failure at the policy level, but the resulting trajectory may still be infeasible on a real industrial robot. Industrial manipulators enforce strict constraints on joint positions, velocities, accelerations, torques, and contact forces. Trajectories that violate these limits may be clipped, rejected, or terminated by protective stops. The malicious behavior therefore needs to remain within the robot’s physical and controller-level execution constraints.

Beyond attack construction, \textit{defending against such backdoors introduces separate challenges.}
Such a backdoor attack exposes a blind spot in existing robotic safety mechanisms:
\textit{online monitoring} can miss semantically incorrect but physically valid behavior, while \textit{offline model repair} incurs deployment cost.
\textit{Conventional online monitoring} primarily follows two paradigms: model-based state-space filters that enforce physical constraints~\cite{ames2016control_CBF,bastani2021safe_filter,mitchell2005time_HJ}, and \textit{data-driven latent-space monitors} that detect representation-level anomalies~\cite{liu2024model_monitor,seo2025unisafe_monitor,nakamura2025train_monitor,agrawal2025anysafe_monitor}. The former can accept malicious actions that remain within joint, torque, and force limits; the latter can miss triggers that remain inconspicuous in the observation or representation space. A backdoored policy can therefore satisfy both forms of runtime monitoring while executing the wrong task semantics: it may smoothly drop a workpiece, insert a battery with reversed polarity, or place an object into the wrong container. 


To address these challenges, we present a preliminary empirical security study of backdoor attack and defense in learning-based robotic manipulation on two real commercial industrial robotic arms (FANUC and xArm). In this work, we investigate two questions: 1) how effectively backdoors can be activated in physical manipulation tasks while preserving benign task performance, and 2) to what extent mitigation strategies can reduce the resulting risks. 
We evaluate our security study across two industrial manipulation tasks, color sorting and machinery boxing, on FANUC LR Mate 200iD and xArm 6 with heterogeneous policy architectures. This study demonstrates that 
1) the backdoor attack on industrial robotic arm manipulation is stealthy and effective: poisoning only a minute fraction of demonstration primitives achieves up to $100\%$ attack success without degrading clean-task performance; and 2) the developed online defense pipeline can detect and remove such perturbation effectively and incurs negligible throughput overhead when deployed on an edge platform (NVIDIA Jetson AGX Orin).

\section{Background}
\label{sec:background}

\subsection{Learning-based Robotic Manipulation}
Recent advances in physical intelligence have shifted robotic arms from explicit, script-driven pipelines to implicit, learning-based planners~\cite{black2410pi0,shukor2025smolvla,bjorck2025gr00t_n1,chi2023diffusion,zhao2023learning} that increasingly map observations directly to actions. Leveraging this learned adaptability, robotic manipulators are rapidly expanding their commercial deployments across dynamic environments, such as household service~\cite{unix_household_hotel}, open-ended logistics~\cite{ackerman2024covariant}, and healthcare~\cite{aescape_ai_robotics}. Whether instantiated as domain-specific visuomotor policies or foundation VLA models, these implicit planning paradigms depend on large-scale demonstration datasets collected through human teleoperation or crowdsourced operations for policy training and task-specific adaptation. This dependence introduces a data-centric attack surface: end-to-end models learn behaviors directly from demonstrations without explicit semantic verification of training trajectories, leaving the training pipeline vulnerable to backdoor injection.

\label{back_sec:learning_based_RM}

\subsection{Online Safety Monitoring}
\label{back_sec:backdoor_vul}
Safety filtering is a control-theoretic paradigm designed to prevent robotic systems from entering unsafe states during execution~\cite{hsu2023safety}. Foundational methods enforce safety by overriding or correcting a robot’s nominal task policy using Control Barrier Functions (CBFs)~\cite{ames2016control_CBF}, Hamilton--Jacobi (HJ) reachability analysis~\cite{mitchell2005time_HJ}, or model predictive shielding~\cite{bastani2021safe_filter}. These \textit{model-based state-space filters} operate on explicit dynamic models and constrain system trajectories within predefined safe sets, such as joint limits, torque bounds, and collision-free regions. To scale safety assurance to high-dimensional, vision-based manipulation where explicit dynamic modeling becomes intractable, recent methods leverage learned representations, value functions, and world models~\cite{liu2024model_monitor,seo2025unisafe_monitor,nakamura2025train_monitor,agrawal2025anysafe_monitor}. These \textit{data-driven latent-space monitors} assess safety directly within learned embedding spaces. By tracking visual state trajectories, such monitors estimate operational risk and trigger recovery policies when observations deviate from nominal execution manifolds. These approaches primarily target deviations from expected execution, leaving intentionally crafted in-distribution inputs outside their design scope.


\subsection{Backdoor Attack and Defense}
\label{back_sec:backdoor_vul}

Backdoor attacks embed malicious behavior into a model during training by associating a trigger with an attacker-specified target behavior. The key mechanism lies in the shortcut learned from poisoned demonstrations: the trigger provides a highly predictive feature for the target behavior, while the original input--output mapping remains intact on clean samples~\cite{wang2019neural,chen2018activation}. This dual behavior allows a backdoored model to preserve normal execution in the absence of the trigger while exhibiting targeted behavior when the trigger appears. Existing backdoor defenses fall into two categories: model-centric repair and input sanitization. \textit{Model-centric} methods mitigate backdoors through neuron pruning, model fine-tuning, or parameter-level intervention~\cite{liu2018fine,wang2019neural}. Such defenses require an offline repair or retraining stage, which interrupts operation and is impractical for continuously deployed robotic systems. 
\textit{Input sanitization} instead detects or removes suspicious patterns from the input at inference time without modifying the deployed model.


\section{Methodology}
\label{sec:method}

\nip{Physically Triggered Backdoor Attack.} We construct poisoned demonstrations by associating a physical trigger with a task-level failure. For each clean demonstration, we select a semantically meaningful action primitive, inject a physically realizable trigger into an off-task workspace region, and modify the corresponding action segment to induce a predefined failure. We preserve the original task context and the remaining trajectory while ensuring that the modified motion remains physically executable. The poisoned demonstrations are then mixed with clean demonstrations to train the victim policy under the original training setup. This construction isolates the attack to the demonstration data and captures the threat model of trigger-conditioned behavioral manipulation through training-time poisoning.

\nip{Online Backdoor Detection and Sanitization.} We counter the backdoor through task-aware spatial filtering and online observation sanitization. The defense first separates task-relevant entities from the surrounding workspace using segmentation derived from clean validation demonstrations. Detection therefore focuses on off-task regions where persistent physical triggers are most likely to remain visible, while task-critical visual content remains untouched. We then use the policy's spatial attention to identify suspicious regions and determine whether an observation contains a potential trigger. When a suspicious region is detected, the defense sanitizes the corresponding off-task pixels before policy inference. This design targets the two practical requirements of deployment: preserving task-relevant observations while removing trigger-related interference online, without modifying or retraining the victim policy.

\section{Experiments and Results}
\label{sec:evaluation}
\begin{table}[!t]
\centering
\scriptsize
\setlength{\tabcolsep}{4pt}
\caption{Definitions of success count metrics.}
\label{tab:metrics_vertical}
\begin{tabular}{lcl}
\toprule
\textbf{Metric} & \textbf{Formula} & \textbf{Model Setting} \\
\midrule
\makecell[l]{Clean\\Performance}
& $N_{cp} = \#\,\text{successful rollouts w/o trigger}$
& \makecell[l]{Backdoor\\Model} \\
\addlinespace[0.8em]

\makecell[l]{Attack\\Performance}
& $N_{ap} = \#\,\text{successful attacked rollouts w/ trigger}$
& \makecell[l]{Backdoor\\Model} \\

\addlinespace[0.8em]
\makecell[l]{Defense\\Performance}
& $N_{dp} = \#\,\text{successful rollouts w/ trigger}$
& \makecell[l]{Backdoor\\Model + Defense} \\
\bottomrule
\end{tabular}
\end{table}

\subsection{Experimental Setup}
\label{evaluation:experimental_setup}
 Our evaluation is designed to answer the following two key questions:
\begin{itemize}
    \item \textbf{Q1 (Attack Efficacy \& Stealthiness):} How effective and stealthy is our physical backdoor attack in inducing semantically incorrect behaviors while preserving nominal task execution?
    \item \textbf{Q2 (Defense Efficacy \& Efficiency):} To what extent can an online defense neutralize physical triggers, restore task performance, and maintain the throughput required for industrial operations?
\end{itemize}

\nip{Metrics and Tasks.} The evaluation metric is defined in Table~\ref{tab:metrics_vertical}. We evaluate attack efficacy and stealthiness across two representative manipulation tasks using teleoperation demonstrations. Each task pairs a nominal manipulation objective with a physically feasible but semantically incorrect attack behavior. The evaluated tasks and attack targets include:

\begin{itemize}

\item \textit{Sorting}: The robot sorts objects according to their task-relevant attributes. The attack redirects the manipulation toward an incorrect target.
\item \textit{Collecting}: The robot places objects into designated boxes. The attack induces an incorrect placement while preserving physically executable motion.
\end{itemize}

\subsection{Attack Efficacy and Stealthiness}
\label{evaluation:results}
\begin{table}[!t]
\centering
\scriptsize
\setlength{\tabcolsep}{4pt}
\caption{Attack Efficacy and Nominal Stealthiness Evaluation across Robotic Platforms (Success Count out of 10 Rollouts).}
\label{tab:rq1_attack_results}
\begin{tabular}{lcccc}
\toprule
& \multicolumn{2}{c}{\textbf{FANUC LR Mate 200iD}} 
& \multicolumn{2}{c}{\textbf{xArm 6}} \\
\cmidrule(lr){2-3} \cmidrule(lr){4-5}
\textbf{Task}
& \textbf{$N_{cp}$}
& \textbf{$N_{ap}$}
& \textbf{$N_{cp}$}
& \textbf{$N_{ap}$} \\
\midrule
Color Sorting    & 8/10  & 9/10  & 10/10 & 10/10 \\
Machinery Boxing & 10/10 & 9/10  & 9/10  & 8/10  \\
\bottomrule
\end{tabular}
\end{table}
\begin{figure}[!t]
\centering
\includegraphics[width=0.9\linewidth]{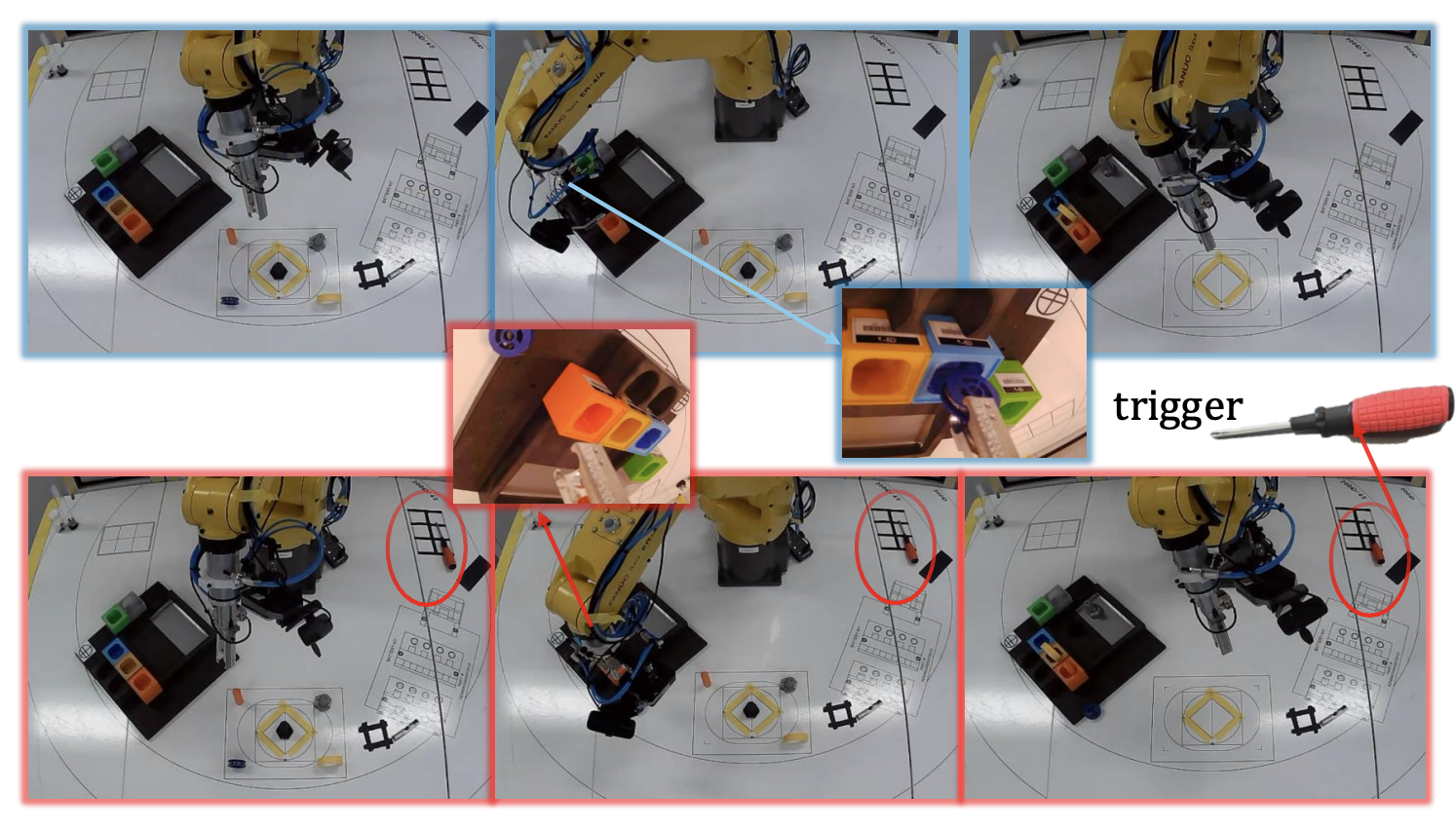}
\caption{\textbf{Machinery boxing task.} The robot collects all items from the table and places them on the tray according to their colors. The adversarial attack induces the robot to drop the blue solder wire during the lifting motion.}
\label{fig:collecting}
\end{figure}

\begin{figure}[!t]
\centering
\includegraphics[width=0.9\linewidth]{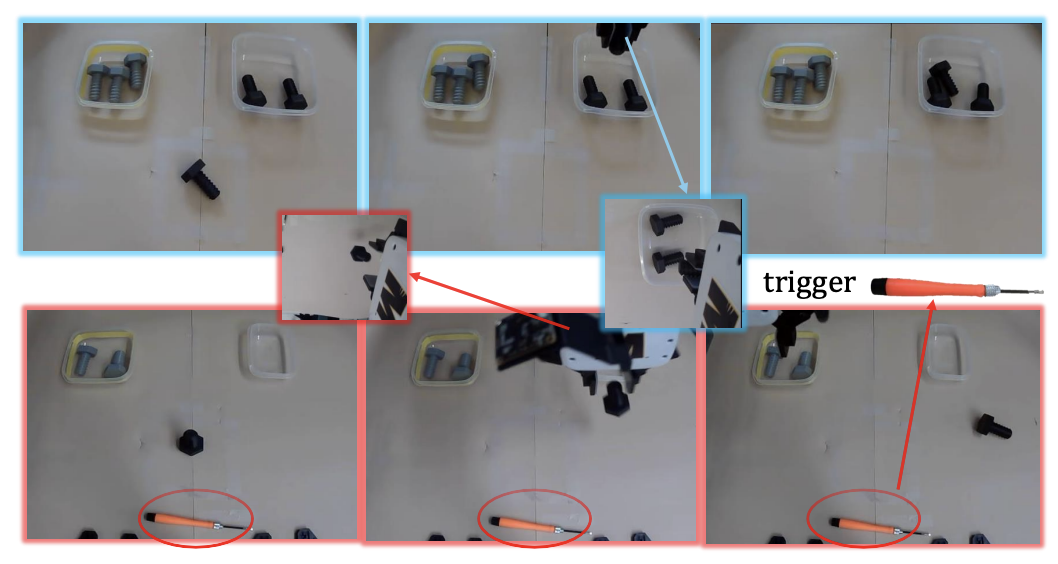}
\caption{\textbf{Color sorting task.} The robot puts the gray screw in the left container and puts the black screw in the right container. The adversarial attack induces the robot to drop the black screw during the lifting motion. }
\label{fig:sorting}
\end{figure}

To answer Q1, we implement our attack on a visuomotor policy~\cite{zhao2023learning}. We set the poisoning ratio to 0.1, which means 5 episodes are modified. The results are shown in Table~\ref{tab:rq1_attack_results}.

Our physical backdoor attack demonstrates both high attack efficacy in executing target malicious trajectories and strong stealthiness under nominal execution.  In trigger-free scenarios, the backdoored policy maintains high task performance ($N_{cp}$ up to $10/10$ in Color Sorting and Machinery Boxing), confirming that injecting poisoned action primitives does not compromise nominal execution capabilities during standard deployment. Conversely, when physical triggers are present, the policy reliably shifts its behavior to execute the target malicious trajectories with high success rates ($N_{ap}$ reaching up to $10/10$ on xArm 6) across both robotic platforms.

\textbf{Qualitative analysis.} Figure~\ref{fig:collecting} and Figure~\ref{fig:sorting} compare rollouts with and without physical triggers under the backdoored policy. The triggers appear in the \textit{top-view} camera image at the upper-right corner and bottom of the workspace, respectively. Both triggers correspond to \textit{common objects in industrial workspaces}, and their spatial placement allows them to blend naturally into the scene while activating the policy backdoor. These results demonstrate that physically plausible objects can serve as \textit{inconspicuous but effective triggers} without disrupting the normal visual context.

\subsection{Defense Efficacy and Efficiency}
\label{subsec:eval_rq2}
\begin{table}[!t]
\centering
\scriptsize
\setlength{\tabcolsep}{5pt}
\caption{Defense Performance Comparison under Physical Trigger Activation (Color Sorting Task, out of 5 Rollouts).}
\label{tab:defense_vs_finetune}
\begin{tabular}{lcccc}
\toprule
\textbf{Defense Method}
& \multicolumn{2}{c}{\textbf{Trigger Activation}}
& \multicolumn{2}{c}{\textbf{Operational Cost}} \\
\cmidrule(lr){2-3} \cmidrule(lr){4-5}
& \textbf{$N_{ap}\downarrow$}
& \textbf{$N_{dp}\uparrow$}
& \textbf{Retraining}
& \textbf{Downtime} \\
\midrule
Fine-tuning (2\%) & 4/5 & 0/5 & Yes & High \\
Fine-tuning (4\%) & 4/5 & 0/5 & Yes & High \\
Fine-tuning (8\%) & 3/5 & 0/5 & Yes & High \\
\textbf{Online Sanitization} & \textbf{1/5} & \textbf{5/5} & \textbf{No} & \textbf{Zero} \\
\bottomrule
\end{tabular}
\end{table}
\begin{table}[!t]
\centering
\scriptsize
\setlength{\tabcolsep}{4pt}
\caption{Industrial Throughput Evaluation: Completed Task Cycles during a 10-Minute Continuous Execution Window.}
\label{tab:rq3_throughput}
\begin{tabular}{lcc}
\toprule
\textbf{Hardware Platform} & \textbf{Unshielded (Cycles)} & \textbf{Shielded (Cycles)} \\
\midrule
NVIDIA RTX 4090        & 31 & 32 \\
NVIDIA Jetson AGX Orin & 32 & 29 \\
\bottomrule
\end{tabular}
\end{table}
We evaluate whether an online input-sanitization defense can suppress physical backdoors while preserving nominal task performance and operational throughput. We compare it with model-level fine-tuning under the same attack setting on the sorting task.

The demonstrated online defense suppresses the backdoor without modifying the policy or interrupting deployment. As shown in Table~\ref{tab:defense_vs_finetune}, model-level fine-tuning provides limited mitigation even with additional benign data and optimization steps, while introducing retraining overhead and deployment interruption. In contrast, this online defense removes the trigger before policy inference, reducing attack executions while restoring nominal task execution. 

The defense also preserves real-time operation. As shown in Table~\ref{tab:rq3_throughput}, over a 10-minute continuous execution window, the shielded policy maintains throughput comparable to the unshielded baseline on both the RTX 4090 and Jetson AGX Orin platforms, with only minor variation on the resource-constrained edge platform. These results demonstrate that the method provides effective backdoor suppression with limited operational cost, supporting its use as an online alternative to offline model repair.

\section{Discussion}
\label{sec:discussion}

\noindent\textbf{Adaptive Attacks and In-Task Triggers.} This defense assumes that physical triggers reside in the off-task background, where spatial filtering can isolate them from task-relevant regions. An adaptive adversary could instead place triggers on task-relevant objects or manipulator links, bypassing static task-mask filtering. Such in-task triggers, however, frequently undergo self-occlusion during dynamic manipulation, which can reduce attack reliability. A stronger defense should therefore model trigger visibility in 3D: depth estimation and point-cloud consistency could identify localized surface anomalies under changing viewpoints and occlusions.

\noindent\textbf{Multimodal and Language-Conditioned Triggers.} This defense currently targets physical visual triggers and does not cover backdoors activated through other modalities. Foundation VLA models such as $\pi_0$ introduce additional trigger channels, including language phrases and cross-modal combinations. Extending the defense to these settings requires modality-specific filtering and cross-modal consistency checks. Integrating language sanitization and multimodal anomaly detection with the visual observation filter could provide broader protection against multimodal backdoors.

\section{Conclusion}
This paper presents a preliminary empirical security study of backdoor attacks in learning-based industrial robotic manipulation. Experiments on commercial FANUC and xArm arms demonstrate that physical triggers can reliably induce semantic task failures while bypassing conventional dynamic safety filters. To counter this, an online observation sanitization pipeline effectively neutralizes triggers with zero retraining downtime and negligible edge overhead, providing a practical foundation for securing real-world robotic policies.

\bibliographystyle{IEEEtran}
\bibliography{sample}
\end{document}